\documentclass{article}
\usepackage{amsmath,amssymb,graphicx,subfig}
\usepackage[usenames,dvipsnames]{color}
\usepackage{algorithm,algorithmic}

\begin{document}

\title{Inference in Hidden Markov Models with Explicit State Duration Distributions}
\author{Michael Dewar, Chris Wiggins, Frank Wood}%


\maketitle

\begin{abstract}
In this letter we borrow from the inference techniques developed for unbounded state-cardinality (nonparametric) variants of the HMM and use them to develop a tuning-parameter free, black-box inference procedure for Explicit-state-duration hidden Markov models (EDHMM). EDHMMs are HMMs that have latent states consisting of both discrete state-indicator and discrete state-duration random variables.  In contrast to the implicit geometric state duration distribution possessed by the standard HMM, EDHMMs allow the direct parameterisation and estimation of per-state duration distributions. As most duration distributions are defined over the positive integers, truncation or other approximations are usually required to perform EDHMM inference.  
\end{abstract}

\section{Introduction}

\label{section}
Hidden Markov models (HMMs) are a fundamental tool for data analysis and exploration.  Many variants of the basic HMM have been developed in response to shortcomings in the original HMM formulation \cite{Rabiner89}.  In this paper we address inference in the explicit state duration HMM (EDHMM).  By state duration we mean the amount of time an HMM dwells in a state.  In the standard HMM specification, a state's duration is implicit and, a priori, distributed geometrically.

The EDHMM  (or, equivalently, the hidden semi-Markov model \cite{Yu10}) was developed to 
allow explicit parameterization and direct inference of state duration 
distributions.   EDHMM estimation and inference can be performed using the 
forward-backward algorithm; though only if the sequence is short or a tight 
``allowable'' duration interval for each state is hard-coded a priori 
\cite{Yu2006}.   If the sequence is short then forward-backward can be run 
on a state representation that allows for all possible durations up to the 
observed sequence length.  If the sequence is long then forward-backward 
only remains computationally tractable if only transitions between durations 
that lie within pre-specified allowable intervals are considered.   If the 
true state durations lie outside those intervals then the resulting model 
estimates will be incorrect: the learned duration distributions can only reflect 
 what is allowed given the pre-specified duration intervals. 

Our contribution is the development of a procedure for EDHMM inference that does not require any hard pre-specification of duration intervals, is efficient in practice, and, as it is an asymptotically exact procedure, does not risk incorrect inference.  The technique we use to do this 
is borrowed from sampling procedures developed for nonparametric  Bayesian HMM variants \cite{vanGael2008}.  Our key insight is simple: the machinery developed for inference in HMMs with a countable number of states is precisely the same as that which is needed for doing inference in an EDHMM with duration distributions over countable support.  So, while the EDHMM is a distinctly parametric model, the tools from nonparametric Bayesian inference can be applied such that black-box inference becomes possible and, in practice, efficient.

In this work we show specifically that a ``beam-sampling'' approach  \cite{vanGael2008} works for  estimating EDHMMs, learning both the transition structure and duration distributions simultaneously.  In demonstrating our EDHMM inference technique we consider a synthetic system  in which the state-cardinality is known and finite, but where each state's duration distribution is unknown. We show that the EDHMM beam sampler performs accurate tracking whilst capturing the duration distributions as well as the probability of transitioning between states.

The remainder of the letter is organised as follows.  In Section~\ref{sec:Model} we introduce the EDHMM; in Section~\ref{sec:inference} we review beam-sampling for the infinite Hidden Markov Model (iHMM) \cite{Beal2002} and show how it relates to the EDHMM inference problem; and in Section~\ref{sec:experiments} we show results from using the EDHMM to model synthetic data.

    \begin{figure}[t]
        \centering
        \subfloat[][]{
            \includegraphics[width=0.5\textwidth]{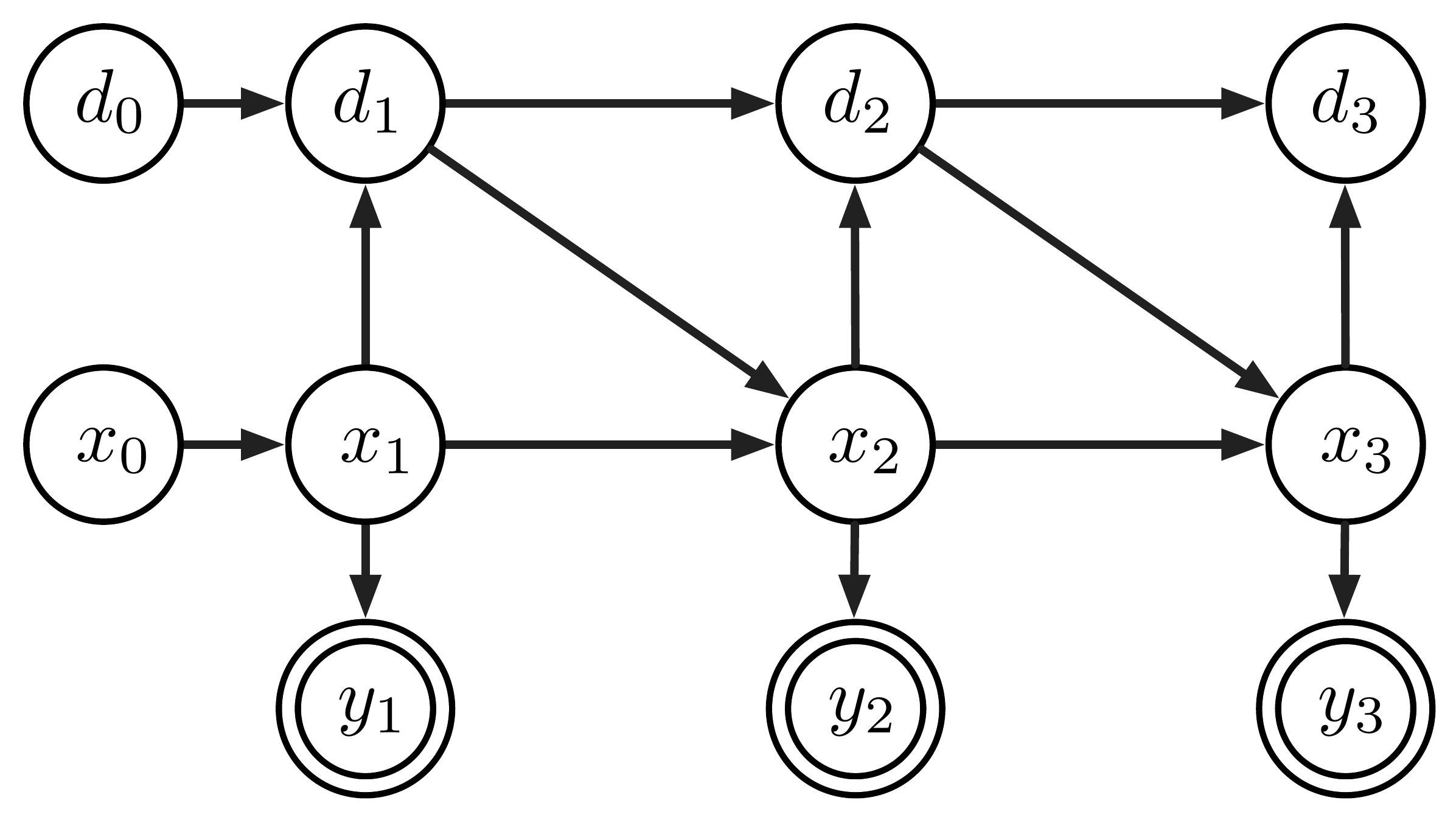}
            \label{fig:graphical model}
        }
        \subfloat[][]{
            \includegraphics[width=0.5\textwidth]{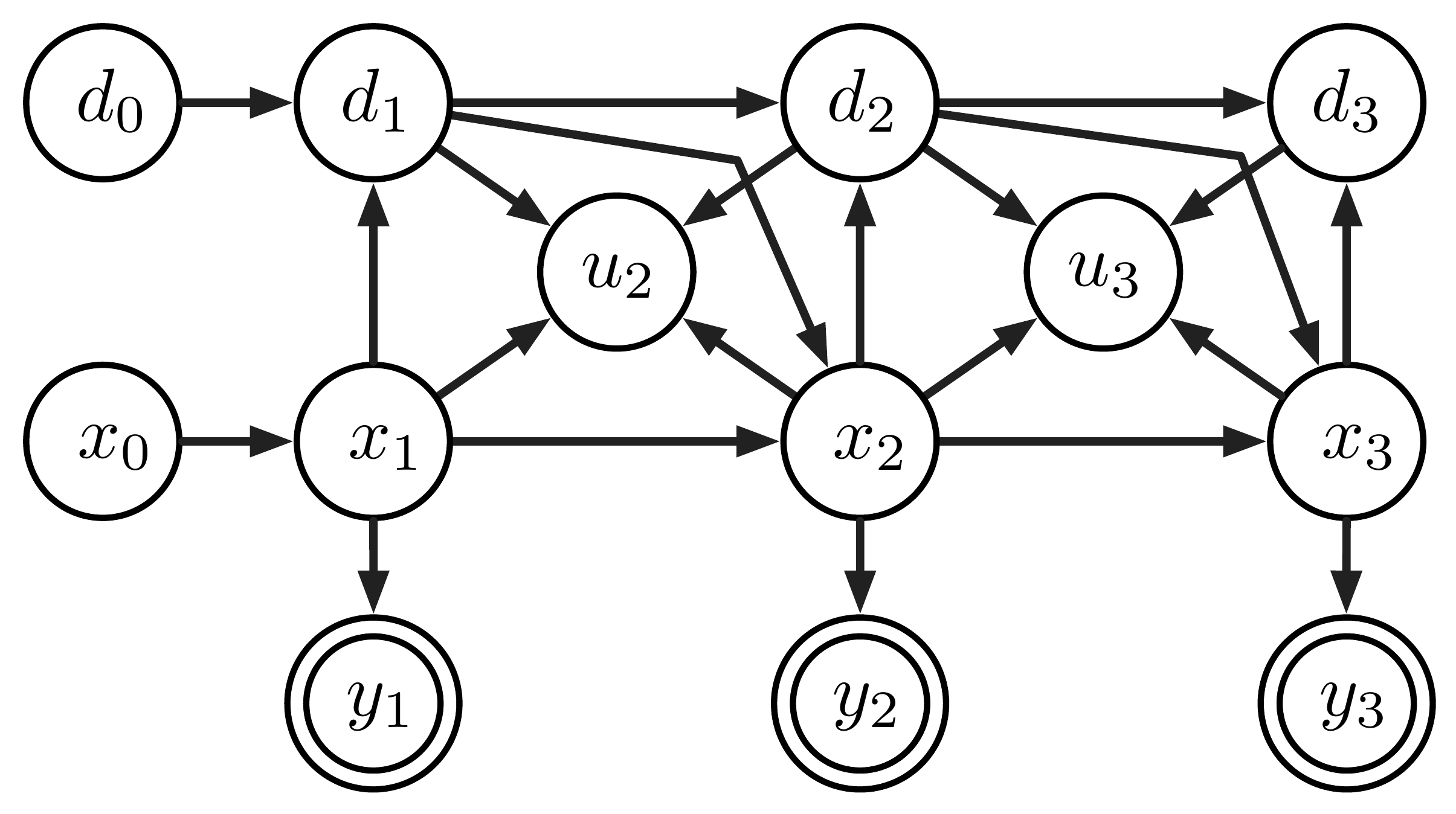}
         \label{fig:aux graphical model}
        }
        \caption{a) The Explicit Duration Hidden Markov Model. The time left in the current state $x_t$ is denoted $d_t$. The observation at each point in time is denoted $y_t$. b) The EDHMM with the additional auxiliary variable $u_t$ used in the beam sampler.}
    \label{fig:graphs}    
    \end{figure}
\section{Explicit Duration Hidden Markov Model}
\label{sec:Model}

The EDHMM captures the relationships among state $x_t$, duration $d_t$, and  observation $y_t$ over time $t$. It consists of four components: the initial state distribution, the transition distributions, the observation distributions, and the duration distributions. 

We define the observation sequence $\mathcal{Y} = \{y_1, y_2, \ldots, y_T\}$; the latent state sequence $\mathcal{X} = \{x_0, x_1, x_2, \ldots, x_T\}$; and the remaining time in each segment $\mathcal{D} = \{ d_1, d_2, \ldots, d_T\}$, where $x_t \in \{ 1, 2, \ldots, K\}$ with $K$ the maximum number of states, $d_t \in \{1, 2, \ldots \}$, and $y_t \in \mathbb{R}^n$.     We assume that the Markov chain on the latent states is homogenous, i.e., that $p(x_t = j | x_{t-1}=i, A) = a_{i,j} \forall t$ where $A$ is a $K\times K$ matrix with element $a_{i,j}$ at row $i$ and column $j.$  The prior on $A$ is row-wise Dirichlet with zero prior mass on self-transitions, i.e.  $p(a_{i,:}) = \mathrm{Dir}({1}/{(K-1)}, \ldots, 0 , \ldots {1}/{K-1})$ where $a_{i,:}$ is a row vector and the $i$th Dirichlet parameter is $0.$  Each state is imbued with its own duration distribution $p(d_t | x_t = k) = p(d_t | \lambda_{k})$ with parameter $\lambda_k$.  Each duration distribution parameter is drawn from a prior $p(\lambda_k)$ which can be chosen in an application specific way.  The collection of all duration distribution parameters is $\lambda = \{\lambda_1, \ldots, \lambda_K\}$.  Each state is also imbued with an observation generating distribution $p(y_t  | x_t = k) = p(y_t | \theta_{k})$ with parameter $\theta_k$.  Each observation distribution parameter is drawn from a prior $p(\theta_k)$ also to be chosen according to the application. The set of all observation distribution parameters is $\theta.$  In the following exposition, explicit conditional dependencies on component distribution parameters are omitted to focus on the particulars unique to the EDHMM.

In an EDHMM the transitions between states are only allowed at the end of a segment:
    \begin{equation}
        p(x_t | x_{t-1}, d_{t-1}) = 
        \begin{cases} 
            \delta(x_t, x_{t-1}) & \textrm{if $d_{t-1} > 1$} \\
            p(x_t | x_{t-1}) & \textrm{otherwise}
        \end{cases}
    \end{equation}
where the Kronecker delta $\delta(a,b) = 1$ if $a=b$ and zero otherwise. The duration distribution generates segment lengths at every state switch:
    \begin{equation}
        p(d_t | x_{t}, d_{t-1}) = 
        \begin{cases} 
            \delta(d_t, d_{t-1}-1) & \textrm{if $d_{t-1} > 1$} \\
            p(d_t | x_{t}) & \textrm{otherwise.}
        \end{cases}
    \end{equation}
The joint distribution of the EDHMM is 
\begin{multline}
    \label{eq:joint}
    p(\mathcal{X},\mathcal{D},\mathcal{Y}) =  
    p(x_0)p(d_0) \\ \prod_{t=1}^T p(y_t | x_t, \theta) p(x_t | x_{t-1}, d_{t-1}, A) p(d_t | x_{t}, d_{t-1}, \lambda)
\end{multline}
corresponding to the graphical model in Figure \ref{fig:graphical model}. Alternative choices to define the duration variable $d_t$ exist; see \cite{Chiappa2011} for details. Algorithm \ref{alg:gen} illustrates the EDHMM as a generative model.

\begin{figure*}[ttt!]
\begin{minipage}[t]{2in}
\begin{algorithm}[H]
    \caption{Generate Data}
    \label{alg:gen}
    \begin{algorithmic}
        \STATE sample $x_0 \sim p(x_0)$, $d_0 \sim p(d_0)$
        \FOR {$t = 1, 2, \ldots, T$}
            \IF{$d_{t-1} = 1$}
                \STATE a new segment starts:
                \STATE sample $x_t \sim p(x_t | x_{t-1})$
                \STATE sample $d_t \sim p(d_t | x_t)$
            \ELSE
                \STATE the segment continues:
                \STATE $x_t = x_{t-1}$
                \STATE $d_t = d_{t-1} - 1$
            \ENDIF
        \STATE sample $y_t \sim p(y_t | x_t)$
        \ENDFOR
    \end{algorithmic}
\end{algorithm} 
 \end{minipage}
 \hfill
 \begin{minipage}[t]{3.3in}
\begin{algorithm}[H]
    \caption{Sample the EDHMM}
    \label{alg:beam}
    \begin{algorithmic}
\STATE Initialise parameters $A$, $\lambda$, $\theta.$ Initialize $u_t$ small $\forall\, T$
\FOR{sweep $ \in \{1,2,3,\ldots \}$}
    \STATE \textbf{Forward}: run \eqref{eqn:scaled forward} to get $\hat{\alpha}_t(z_t)$ given $\mathcal{U}$ and $\mathcal{Y} \; \forall\, T$
    \STATE \textbf{Backward}: sample $z_T \sim \hat{\alpha}_T(z_T)$
    \FOR{$t \in \{T, T-1, \ldots, 1\}$}
        \STATE sample $z_{t-1} \sim \mathbb{I}(u_t < p(z_{t} | z_{t-1}))\hat{\alpha}_{t-1}(z_{t-1})$
    \ENDFOR
\STATE \textbf{Slice:}\FOR {$t \in \{1, 2, \ldots, T \}$}
    \STATE evaluate $l = p(d_t|x_t,d_{t-1})p(x_t|x_{t-1},d_{t-1})$
    \STATE sample $u_{t} \sim \mathrm{Uniform}(0,l)$
    \ENDFOR
\STATE sample parameters $A$, $\lambda$, $\theta$
\ENDFOR
\end{algorithmic}
\end{algorithm}
\end{minipage}
 \hfill
\end{figure*}

\section{EDHMM Inference}

\label{sec:inference}
Our aim is to estimate the conditional posterior distribution of the latent states ($\mathcal{X}$ and $\mathcal{D}$) and parameters ($\theta, \lambda$ and $A$) given observations $\mathcal{Y}$ by samples drawn via Markov chain Monte Carlo. Sampling $\theta$ and $A$ given $\mathcal{X}$ proceeds per usual textbook approaches \cite{Bishop06}.  Sampling $\lambda$ given $\mathcal{D}$ is straightforward in most situations.  Indirect Gibbs sampling  of $\mathcal{X}$ is possible using auxiliary state-change indicator variables, but for reasons similar to those in \cite{Goldwater2009}, such a sampler will not mix well.  The main contribution of this paper is to show how to generate posterior samples of  $\mathcal{X}$ and $\mathcal{D}$.

\subsection{Forward Filtering, Backward Sampling}

We can, in theory, use the forward messages from the forward backward algorithm \cite{Rabiner89} to sample the conditional posterior distribution of $\mathcal{X}$ and $\mathcal{D}.$   To do this we treat each state-duration tuple as a single random variable 
(introducing the notation $z_t = \{x_t,d_t\}$).  
Doing so recovers the standard hidden Markov model structure and hence standard forward messages can be used directly.  A forward filtering, backward sampler for $\mathcal{Z} = \{z_1, \ldots, z_T\}$ conditioned on 
all other random variables
requires the classical forward messages:
    \begin{equation}
        \alpha_t(z_t) = 
        \sum_{z_{t-1}}
        p(z_t | z_{t-1}) 
        p(y_t|z_t) 
        \alpha_{t-1}(z_{t-1}) 
        \label{eqn:forward recursion}
    \end{equation}
     where the transition probability can be factorised according to our modelling assumptions:
     \begin{equation}
        p(z_{t} | z_{t-1}) = p(x_t | x_{t-1}, d_{t-1}) p(d_t | d_{t-1}, x_t).
     \end{equation}

Unfortunately the sum in \eqref{eqn:forward recursion} has at worst an infinite number of terms in the case of  duration distributions with countably infinite support and at best a very large number of terms in the case of long sequences. The standard approach to EDHMM inference involves truncating considered durations to only those that lie between $d_\mathrm{min}$ and $d_\mathrm{max}$ or computation involving all possible durations up to the observed length of the sequence ($d_\mathrm{min}=0, d_\mathrm{max}=T$). This leads to per-sample, forward-backward computational complexity of $O(T(K(d_\mathrm{max}-d_\mathrm{min}))^2)$.
Truncation yields inference that will simply fail if an actual duration lies outside hard-coded allowable duration intervals.   Considering all possible durations up to length $T$ is often computationally impossible.
The beam-sampler we propose behaves like a dynamic version of the truncation approach, automatically defining and scaling per-state duration truncation intervals.   Better though, the way it does this results in an asymptotically exact sample with no risk of incorrect inference resulting from incorrectly pre-specified duration truncations.   We do not characterize the computational complexity of the proposed beam sampler in this work but note that it is upper bounded by $O(T(KT)^2)$ (i.e., the beam sampler admits durations of length equal to the entire sequence) but in practice is found to be as  or more efficient than the risky hard-truncation approach.


\subsection{EDHMM Beam Sampling}

A recent contribution to inference 
in 
the infinite Hidden Markov Model (iHMM) \cite{Beal2002} suggests a way around truncation \cite{vanGael2008}.  The iHMM is an HMM with a countable number of states.  Computing the forward message for a forward filtering, backward sampler for the latent states in an iHMM also requires a sum over a countable number of elements.  
The ``beam sampling'' approach  \cite{vanGael2008}, which we can apply largely without modification, is to truncate this sum by introducing a ``slice'' \cite{Neal2003} auxiliary variable $\mathcal{U} = \{u_1, u_2, \ldots,u_T\}$ at each time step.  The auxiliary variables are chosen in such a way as to automatically limit each sum in the forward pass to a finite number of terms while still allowing all possible durations.


The particular choice of auxiliary variable $u_t$ is important.  We follow  \cite{vanGael2008} in choosing $u_t$ to be conditionally distributed given the current and previous state and duration in the following way (see the graphical model in Figure \ref{fig:aux graphical model}):
\begin{equation}
    \label{eqn:slice}
    p(u_t | z_t, z_{t-1}) = 
    \frac
    {\mathbb{I}(0 < u_t < p(z_t | z_{t-1}))} 
    {p(z_t | z_{t-1})}
\end{equation}
where $\mathbb{I}(\cdot)$ returns one if its operand is true and zero otherwise. Given $\mathcal{U}$ it is possible to sample the state $\mathcal{X}$ and duration $\mathcal{D}$ conditional posterior. 
Using notation $\mathcal{Y}_{t_1}^{t_2} = \{y_{t_1}, y_{t_1+1}, \ldots,y_{t_2}\}$  to indicate sub-ranges of a sequence, the new forward messages we compute are:
\begin{eqnarray}
   \hat{\alpha}_t(z_t) &=& 
   p(z_t, \mathcal{Y}_1^t , \mathcal{U}_1^{t})   \label{eqn:scaled forward} 
   =
   \sum_{z_{t-1}}
   p(z_t, z_{t-1} , \mathcal{Y}_1^t , \mathcal{U}_1^{t}) \\ 
   &\propto& 
   \sum_{z_{t-1}}
   p(u_{t} | z_t, z_{t-1})
   p(z_t, z_{t-1} , \mathcal{Y}_1^t, \mathcal{U}_1^{t-1}) \nonumber \\
   &=& 
   \sum_{z_{t-1}}
   \mathbb{I}(0 < u_{t} < p(z_t | z_{t-1}))
   p(y_t|z_t) \hat{\alpha}_{t-1}(z_{t-1}) \nonumber.
\end{eqnarray}
The indicator function $\mathbb{I}$ results in non-zero probabilities in the forward message for only those states $z_t$ whose likelihood given $z_{t-1}$ is greater than $u_t$. The beam sampler derives its computational advantage from the fact that the set of $z_t$'s for which this is true is typically small. 


The backwards sampling step recursively samples a state sequence from the distribution $p(z_{t-1} | z_{t}, \mathcal{Y}, \mathcal{U})$
which can expressed in terms of the forward variable:
\begin{eqnarray}
    p(z_{t-1} | z_{t}, \mathcal{Y}, \mathcal{U}) &\propto& p(z_{t},z_{t-1}, \mathcal{Y}, \mathcal{U})  \label{eqn:backward} \\
    & \propto &  
        p(u_{t} | z_t, z_{t-1})p(z_{t}|z_{t-1}) \hat{\alpha}_{t-1}(z_{t-1})
        \nonumber\\
    & \propto & 
       \mathbb{I}(0 < u_{t} < p(z_{t} | z_{t-1}))
        \hat{\alpha}_{t-1}(z_{t-1}).\nonumber
\end{eqnarray}


The full EDHMM beam sampler is given in Algorithm \ref{alg:beam}, which makes use of the forward recursion in \eqref{eqn:scaled forward}, the slice sampler in \eqref{eqn:slice}, and the backwards sampler in \eqref{eqn:backward}.

\subsection{Related Work}

The need to accommodate explicit state duration distributions in HMMs has long been recognised. Rabiner \cite{Rabiner89} details the basic approach which expands the state space to include dwell time before applying a slightly modified Baum-Welch algorithm. This approach specifies a maximum state duration, limiting practical application 
to cases with short sequences and dwell times.
This approach, generalised under the name ``segmental hidden Markov models'', includes more general transitions than those Rabiner considered, allowing the next state and duration to be conditioned on the previous state and duration \cite{Gales93}. Efficient approximate inference procedures were developed in the context of speech recognition \cite{Ostendorf96}, speech synthesis \cite{Zen07}, and evolved into symmetric approaches suitable for practical implementation \cite{Yu2006}. Recently, a ``sticky'' variant of the hierarchical Dirichlet process HMM (HDP-HMM) has been developed \cite{Fox2008}.  The HDP-HMM has countable state-cardinality \cite{Teh06} allowing estimation of the number of states in the HMM; the sticky aspect addresses long dwell times by introducing a parameter in the prior that favours self-transition.

\section{Experiments}

\label{sec:experiments}

\subsection{Synthetic Data}

\begin{figure}
    \subfloat[][]{
        \includegraphics[width=\textwidth]{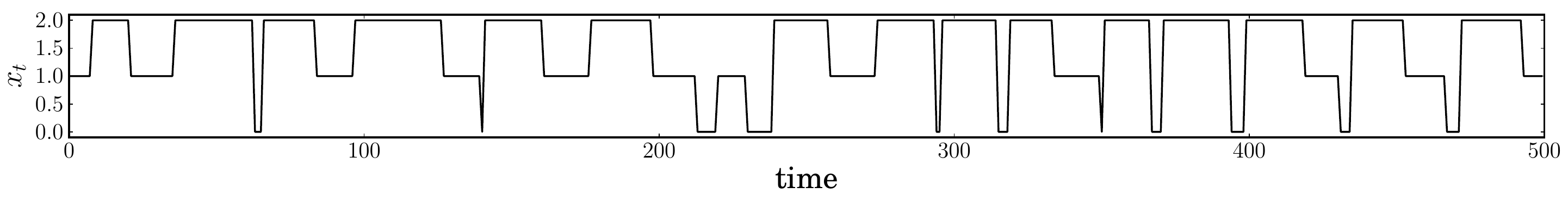}
        \label{fig:exp_1_state}
    } \\
    \subfloat[][]{
        \includegraphics[width=\textwidth]{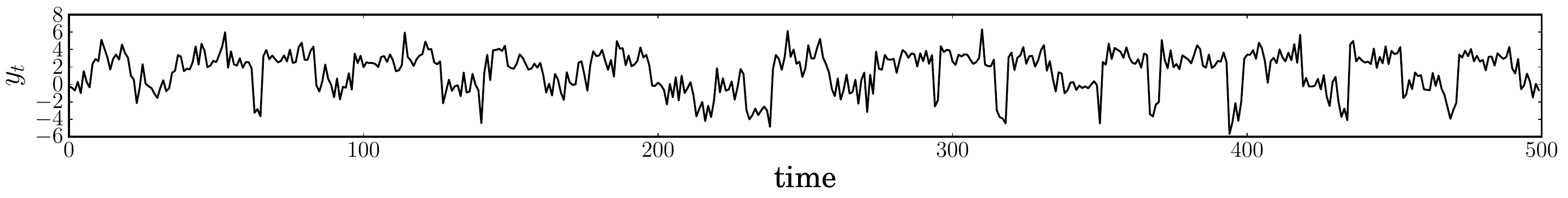}
        \label{fig:exp_1_data}
    }
    \caption{Example a) state and b) observation sequence generated by the explicit duration HMM. Here $K$ = 3; $p(y_t|x_t=j) = \mathrm{N}(\mu_j, 1)$ with $\mu_1 = -3$, $\mu_2 = 0$, and $\mu_3 = 3$; and $p(d_t|x_t=j) = \mathrm{Poisson}(\lambda_j)$ with $\lambda_1 = 5$, $\lambda_2 = 15$, and $\lambda_3 = 20$.}
    \label{fig:experiment1_data}
\end{figure}

The first experiment uses the 500 data points (Figure \ref{fig:experiment1_data}) generated from a three state EDHMM. The duration distributions were Poisson with rates $\lambda_1 = 5$, $\lambda_2 = 15$, $\lambda_3 = 20$; each observation distribution was Gaussian with means of $\mu_1 = -3$, $\mu_2 = 0$, and $\mu_3 = 3$, each with a variance of 1. The transition distributions $A$ were set to
\begin{equation*}
\begin{bmatrix}
    0 & 0.3 & 0.7 \\ 0.6 & 0 & 0.4 \\ 0.3 & 0.7 & 0
\end{bmatrix}.  
\end{equation*}

Broad, uninformative priors were chosen for the parameters of the duration and observation distributions. The observation distribution parameters were given a normal-inverse-Wishart (N-IW) prior with parameters $\nu_0 = 2$, $\Lambda_0 = 1$, $\kappa=0.1$ and $\mu_0 = 0$. The rate parameters for all states were given $\mathrm{Gamma}(1, 10^{5})$ priors. 

\begin{figure}
    \subfloat[][]{
        \includegraphics[width=0.5\textwidth]{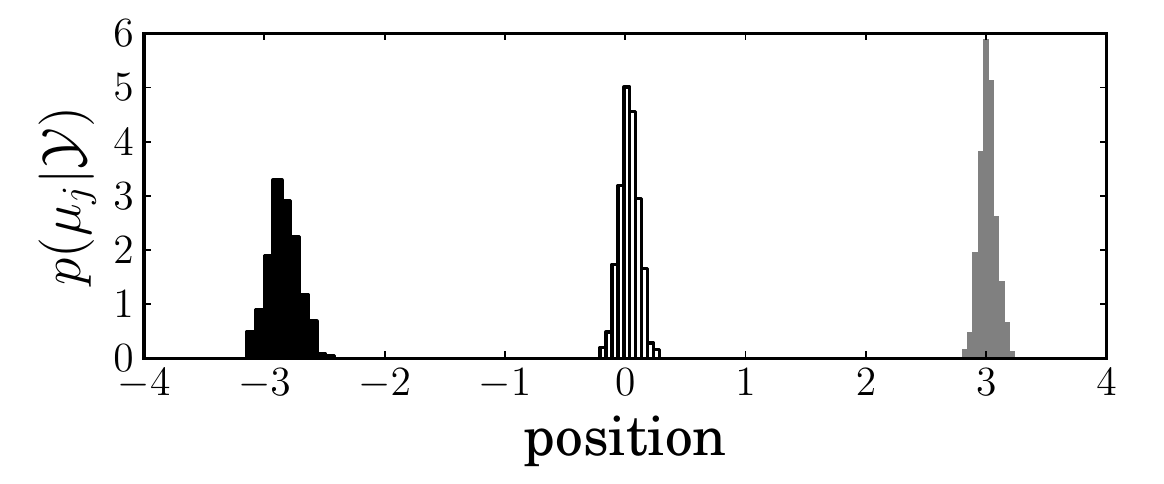}
        \label{fig:posterior_means}
    } 
    \subfloat[][]{
        \includegraphics[width=0.5\textwidth]{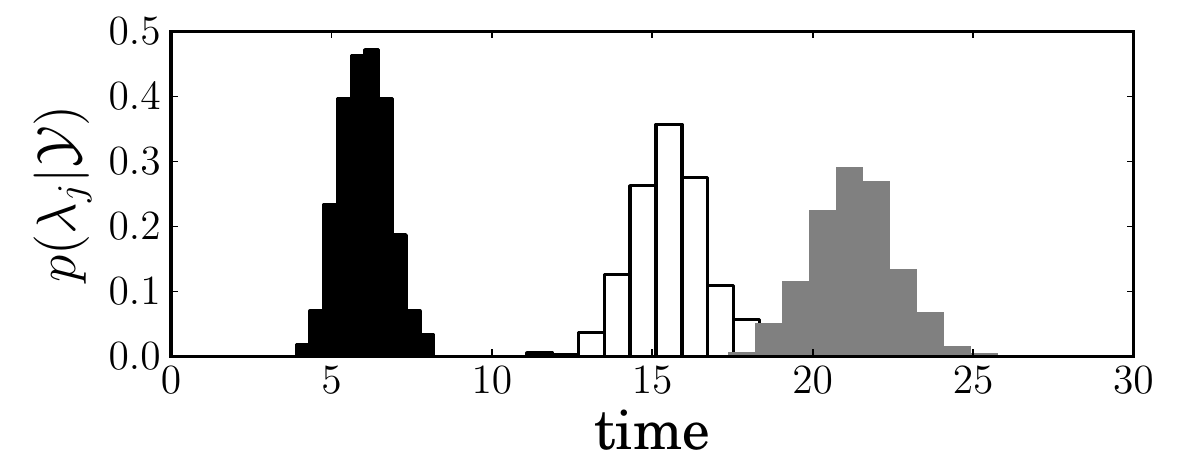}
        \label{fig:posterior_rates}
    }
    \caption{Samples from the posterior distributions of a) the observation distribution means and b) the duration distribution rate parameters for the data shown in Figure \ref{fig:experiment1_data}.}
    \label{fig:experiment1_results}
\end{figure}

\begin{figure}
    \includegraphics[width=\textwidth]{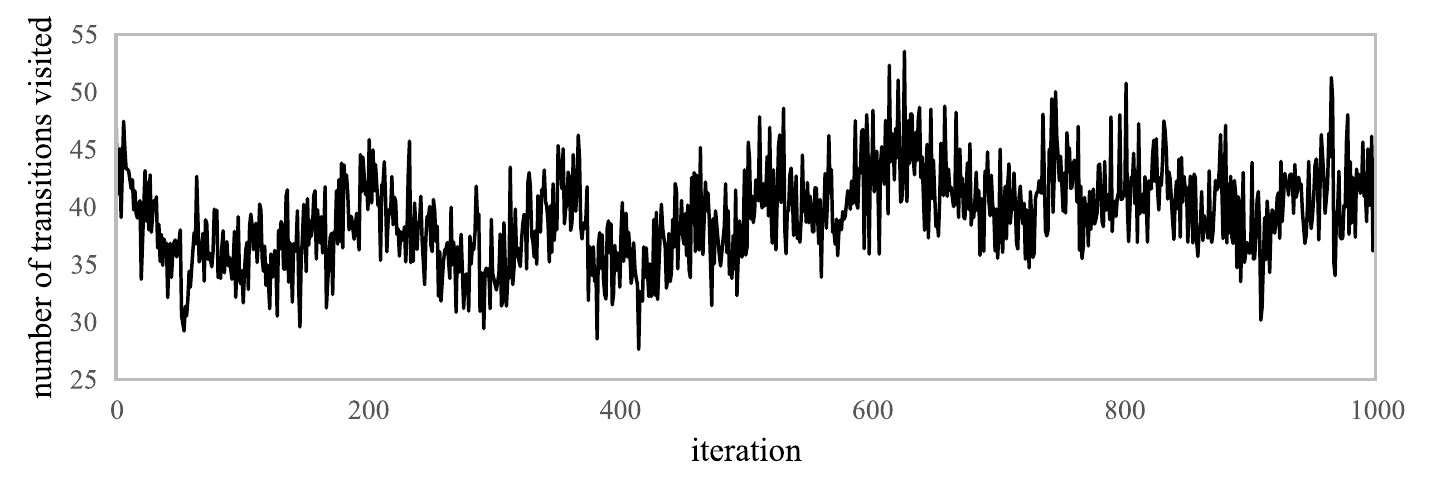}

\caption{Mean number of transitions considered per time point by the beam sampler for 1000 post-burn-in sweeps on data from Figure \ref{fig:experiment1_results}. Consider this in comparison to the $(KT)^2 = O(10^6)$ per time point transitions that would need to be considered by standard forward backward without truncation, a surely-safe, truncation-free, but computationally impractical alternative.}
\label{fig:allowed}
\end{figure}

One thousand samples were collected from the EDHMM beam sampler after a burn-in of 500 samples. The learned posterior distribution of the state duration parameters and means of the observation distributions are shown in Figure \ref{fig:experiment1_results}.  The EDHMM achieves high accuracy in the estimated posterior distribution of the observation means, despite the overlap in observation distributions. The rate parameter distributions are reasonably estimated given the small number of observed segments. Figure \ref{fig:allowed} shows the mean number of transitions visited per time point over each iteration of the sampler. 

\begin{figure}
    \subfloat[][]{
        \includegraphics[width=\textwidth]{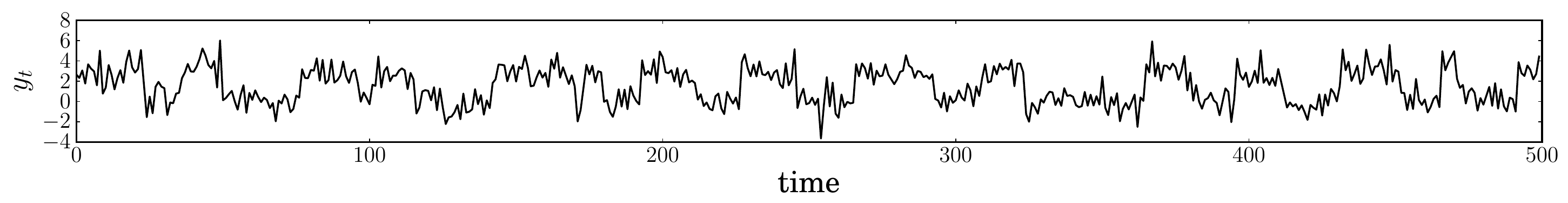}
        \label{fig:exp_2_data}
    } \\
    \subfloat[][]{
        \includegraphics[width=\textwidth]{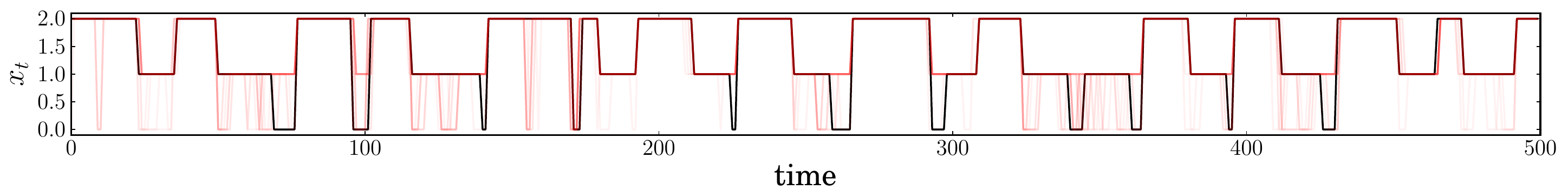}
        \label{fig:exp_2_state}
    } \\
    \subfloat[][]{
        \includegraphics[width=0.5\textwidth]{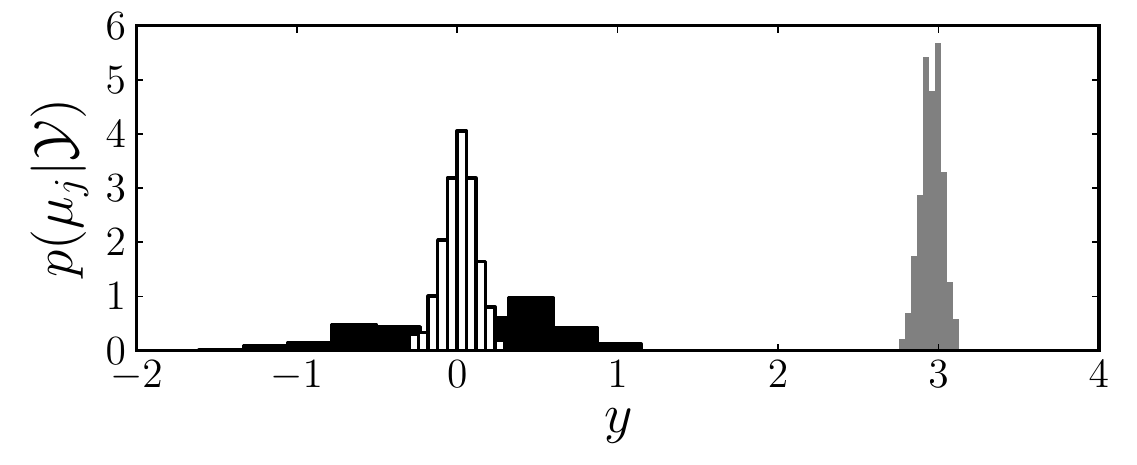}
        \label{fig:posterior_means_3}
    } 
    \subfloat[][]{
        \includegraphics[width=0.5\textwidth]{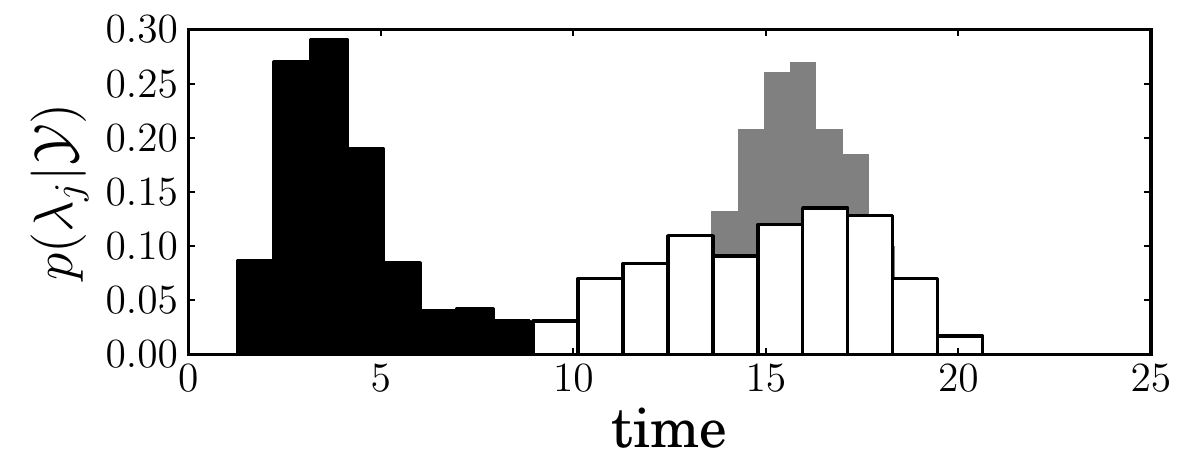}
        \label{fig:posterior_rates_3}
    }
    \caption{Beam sampler results from a system with identical observation distributions but differing durations. Observations are shown in a); true states in b) overlaid with 20 state traces produced by the sampler. Here we have parameters $\mu_1 = \mu_2 = 0$, $\mu_3 = 3$ and $\lambda_1 = 5$, $\lambda_2 = 15$, $\lambda_3 = 20$. Samples from the posterior observation-mean and duration-rate distributions are shown in c) and d), respectively.}
    \label{fig:experiment2_results}
\end{figure}

A second experiment was performed to demonstrate the ability of the EDHMM to distinguish between states having differing duration distributions but the same observation distribution. The same model and sampling procedure was used as above except here $\mu_1 = 0$, $\mu_2 = 0$, and $\mu_3 = 3$.
Figure~\ref{fig:experiment2_results} shows that the sampler clearly separates the high state associated with $\mu_3$ from the other states and clearly 
reveals the presence of
two  low states with differing duration distributions. Figure~\ref{fig:exp_2_state} shows posterior samples that indicate that the model is mixing over ambiguities about states $0$ and $1$ as it should. 


\section{Discussion}

\label{sec:dicussion}

We presented a beam sampler for the explicit state duration HMM. This sampler draws state sequences from the true posterior distribution without any need to make truncation approximations.
It remains future work to combine the explicit state duration HMM and the iHMM.
Python code associated with the EDHMM is available online.\footnote{http://github.com/mikedewar/EDHMM}


\bibliographystyle{plain}

\bibliography{library}

\begin{thebibliography}{10}

\bibitem{Beal2002}
Matthew~J Beal, Z~Ghahramani, and C~E Rasmussen.
\newblock {The Infinite Hidden Markov Model}.
\newblock {\em Advances in Neural Information Processing Systems}, 1:577--584,
  2002.

\bibitem{Bishop06}
C~M Bishop.
\newblock {\em {Pattern Recognition and Machine Learning}}.
\newblock Springer, 2006.

\bibitem{Chiappa2011}
Silvia Chiappa.
\newblock {Unified Treatment of Hidden Markov Switching Models}, April 2011.

\bibitem{Fox2008}
Emily~B Fox, Erik~B Sudderth, Michael~I Jordan, and Alan~S Willsky.
\newblock {An HDP-HMM for systems with state persistence}.
\newblock {\em Proceedings of the 25th International Conference on Machine
  Learning (2008)}, 25:312--319, 2008.

\bibitem{Gales93}
M~J~F Gales and S~J Young.
\newblock {The Theory of Segmental Hidden Markov Models}.
\newblock Technical report, Cambride University Engineering Department, 1993.

\bibitem{Goldwater2009}
S.~Goldwater, T.L. Griffiths, and M.~Johnson.
\newblock {A Bayesian framework for word segmentation: Exploring the effects of
  context}.
\newblock {\em Cognition}, 112(1):21--54, 2009.

\bibitem{Neal2003}
Radford~M Neal.
\newblock {Slice sampling}.
\newblock {\em Annals of Statistics}, 31(3):705--767, 2003.

\bibitem{Ostendorf96}
M~Ostendorf, V~V Digalakis, and O~A Kimball.
\newblock From {HMM}s to segment models: a unified view of stochastic modeling
  for speech recognition.
\newblock {\em IEEE Transactions on Speech and Audio Processing}, 4(5):360 --
  378, 1996.

\bibitem{Rabiner89}
L~R Rabiner.
\newblock {A Tutorial on Hidden Markov Models and Selected Applications in
  Speech Recognition}.
\newblock {\em Proceedings of the IEEE}, 77(2):257--286, 1989.

\bibitem{Teh06}
Y~W Teh, M~I Jordan, M~J Beal, and D~M Blei.
\newblock {Hierarchical Dirichlet Processes}.
\newblock {\em Journal of the American Statistical Association},
  101(476):1566--1581, 2006.

\bibitem{vanGael2008}
Jurgen {Van Gael}, Yunus Saatci, Yee~Whye Teh, and Zoubin Ghahramani.
\newblock {Beam sampling for the infinite hidden Markov model}.
\newblock {\em Proceedings of the 25th International Conference on Machine
  Learning (2008)}, 25:1088--1095, 2008.

\bibitem{Yu10}
S.~Yu.
\newblock Hidden semi-{M}arkov models.
\newblock {\em Artificial Intelligence}, 174:215--243, 2010.

\bibitem{Yu2006}
Hisahi {Yu, Shun-Zeng, Kobayashi}.
\newblock {Practical implementation of an efficient forward-backward algorithm
  for an explicit-duration hidden Markov model}.
\newblock {\em IEEE Transactions on Signal Processing}, 54(5):1947--1951, 2006.

\bibitem{Zen07}
Heiga Zen, Keiichi Tokuda, Takashi Masuko, Takao Kobayasih, and Tadashi
  Kitamura.
\newblock A hidden semi-{M}arkov model-based speech synthesis system.
\newblock {\em IEICE Transactions on Information and Systems},
  E90-D(5):825--834, 2007.

\end{thebibliography}

\end{document}